\documentclass[sigconf]{acmart}
\AtBeginDocument{%
  }

\setcopyright{acmlicensed}
\copyrightyear{2026}
\acmYear{2026}
\setcopyright{none}
\acmConference[ICMR '26]{International Conference on Multimedia Retrieval}{June 16--19, 2026}{Amsterdam, Netherlands}
\acmBooktitle{International Conference on Multimedia Retrieval (ICMR '26), June 16--19, 2026, Amsterdam, Netherlands}
\acmDOI{10.1145/3805622.3810849}
\acmISBN{979-8-4007-2617-0/2026/06}

\usepackage{subcaption}
\usepackage{multirow}
\usepackage{enumitem}

\begin{document}

\title{AutoAWG: Adverse Weather Generation with Adaptive Multi-Controls for Automotive Videos}

\author{Jiagao Hu}
\affiliation{%
  \institution{MiLM Plus, Xiaomi Inc.}
  \city{Wuhan}
  \country{China}
}
\email{hujiagao@xiaomi.com}

\author{Daiguo Zhou}
\affiliation{%
  \institution{MiLM Plus, Xiaomi Inc.}
  \city{Wuhan}
  \country{China}
}
\email{zhoudaiguo@xiaomi.com}

\author{Danzhen Fu}
\affiliation{%
  \institution{MiLM Plus, Xiaomi Inc.}
  \city{Wuhan}
  \country{China}
}
\email{fudanzhen@xiaomi.com}

\author{Fuhao Li}
\affiliation{%
  \institution{MiLM Plus, Xiaomi Inc.}
  \city{Wuhan}
  \country{China}
}

\author{Zepeng Wang}
\affiliation{%
  \institution{MiLM Plus, Xiaomi Inc.}
  \city{Wuhan}
  \country{China}
}

\author{Fei Wang}
\affiliation{%
  \institution{MiLM Plus, Xiaomi Inc.}
  \city{Wuhan}
  \country{China}
}

\author{Wenhua Liao}
\affiliation{%
  \institution{MiLM Plus, Xiaomi Inc.}
  \city{Wuhan}
  \country{China}
}

\author{Jiayi Xie}
\affiliation{%
  \institution{MiLM Plus, Xiaomi Inc.}
  \city{Wuhan}
  \country{China}
}

\author{Haiyang Sun}
\affiliation{%
  \institution{Xiaomi Inc.}
  \city{Shanghai}
  \country{China}
}

\renewcommand{\shortauthors}{Hu et al.}

\begin{abstract}

Perception robustness under adverse weather remains a critical challenge for autonomous driving, with the core bottleneck being the scarcity of real-world video data in adverse weather. Existing weather generation approaches struggle to balance visual quality and annotation reusability. We present \emph{AutoAWG}, a controllable \emph{A}dverse \emph{W}eather video \emph{G}eneration framework for \emph{Auto}nomous driving. Our method employs a semantics-guided adaptive fusion of multiple controls to balance strong weather stylization with high-fidelity preservation of safety-critical targets; leverages a vanishing point-anchored temporal synthesis strategy to construct training sequences from static images, thereby reducing reliance on synthetic data; and adopts masked training to enhance long-horizon generation stability. On the nuScenes validation set, \emph{AutoAWG} significantly outperforms prior state-of-the-art methods: without first-frame conditioning, FID and FVD are relatively reduced by 50.0\% and 16.1\%; with first-frame conditioning, they are further reduced by 8.7\% and 7.2\%, respectively. Extensive qualitative and quantitative results demonstrate advantages in style fidelity, temporal consistency, and semantic--structural integrity, underscoring the practical value of \emph{AutoAWG} for improving downstream perception in autonomous driving. Our code is available at: \url{https://github.com/higherhu/AutoAWG}

\end{abstract}



\maketitle

\newcommand{\method}{\emph{AutoAWG}}

\section{Introduction}
Perception robustness under adverse weather (e.g., nighttime, rain, snow, fog) is a key challenge for autonomous driving, and its core bottleneck lies in the extreme scarcity of real-world data~\citep{zhang2023perception}. Among existing solutions, weather removal methods~\citep{ni2021controlling, valanarasu2022transweather, yang2024genuine} are difficult to deploy due to real-time constraints, while weather generation methods~\citep{li2022weather, lan2024sustechgan, zhao2024revisiting} often fail to preserve the original scene structure, making annotations non-reusable and thus costly. Video weather style transfer offers a practical alternative: it can synthesize diverse weather conditions while maximally preserving original annotations, providing efficient and low-cost data augmentation for perception models. However, this task imposes dual stringent requirements: it must produce highly realistic weather appearances and simultaneously preserve the geometry and semantics of safety-critical objects.

Based on this, we argue that an effective model for autonomous-driving video weather style transfer should possess two core capabilities: (1) \textbf{style fidelity and temporal consistency} — faithfully reproducing the visual characteristics of the target weather and maintaining consistent style across consecutive frames; and (2) \textbf{semantic-structural consistency} — precisely preserving the semantics and geometry of safety-critical objects (vehicles, pedestrians, traffic signs) under domain shifts, over time, and across multi-camera views, thereby ensuring that the translated videos remain usable for downstream perception tasks.

To enforce semantic–structural consistency, {prior work commonly introduces high-level controls such as 3D bounding boxes, BEV maps, or trajectory maps~\citep{gao2023magicdrive, gao2024vista, wang2024driving, zhao2024drivedreamer2, XieLWC025}}. While these representations help ensure multi-view and temporal geometric consistency, they lack fine-grained guidance for textures and local structures, leading to limited visual detail and under-expressive or less realistic weather effects (e.g., overly bright night scenes). On the other hand, to secure style fidelity and consistency, many approaches rely heavily on paired data for supervision~\citep{zhou2024simgen, lin2025controllable, song2025omniconsistency}. However, acquiring such paired data of the same scene under multiple weather conditions is practically infeasible in real world. As a workaround, synthetic data are often used for training, but they introduce non-trivial domain gaps, and the synthesis pipeline itself is costly and prone to artifacts and geometric/texture biases.

To address these limitations from a structure–style decoupling perspective, we propose \textbf{\method}. Our key insight is to introduce a set of complementary control conditions and to conceptualize their fusion as a ``coloring-book” process, which tackles the challenges of control granularity and data dependence. Concretely, we formulate video weather transfer as generation guided by structural priors: Lineart outlines objects' boundaries and shapes; Depth and Sketch jointly define the global scene structure and layering; and a semantic segmentation mask partitions the canvas into distinct coloring regions. Building on this foundation, our semantics-guided adaptive fusion and importance-weighted loss effectively bold the contours for safety-critical regions (e.g., vehicles, pedestrians). Under such constraints, the diffusion model is simplified to filling appropriate colors and textures for each region according to a target-weather palette. This explicit disentanglement of style (coloring) from content (structural sketch) ensures strong stylization while preserving object integrity.

To address data scarcity, we further propose a vanishing point-anchored temporal synthesis: by keeping the normalized location of the vanishing point fixed, we generate an equal-ratio cropping sequence, resize all crops to a common resolution, and concatenate them along the temporal dimension to form a pseudo-video that simulates stable forward motion from still images. This substantially mitigates the scarcity of adverse-weather videos and the domain gaps introduced by synthetic pipelines. Finally, to support long-horizon video generation, we adopt a masked segmented training strategy. By either randomly masking all frames or keeping only the first frame while masking the rest, the model is compelled to learn long-range temporal dependencies, ensuring indefinite continuation and temporal consistency in the generated videos.

\textbf{Core Advantages of \method\ : Three Highs}

\begin{itemize}
  \item \textbf{High-Quality Generation.} With semantics-guided adaptive fusion of multiple controls and importance weighting, the model dynamically allocates controls to different regions, striking a balance between strong weather stylization and high-fidelity objects preservation.
  \item \textbf{High Consistency and Reusability.} By directly leveraging pixel-level controls extracted from the input videos, \method\ keeps object geometry and semantics tightly aligned with the original scene. The translated videos can therefore reuse existing annotations (e.g., 2D/3D labels, LiDAR) without re-annotation, enabling plug-and-play integration into downstream tasks.
  \item \textbf{High Flexibility and Scalability.} Thanks to robust controllability and training strategies, the framework naturally supports multi-camera systems and arbitrary-length sequences, meeting practical requirements while reducing cross-frame fluctuation.
\end{itemize}

\begin{figure*}
  \includegraphics[width=0.9\textwidth]{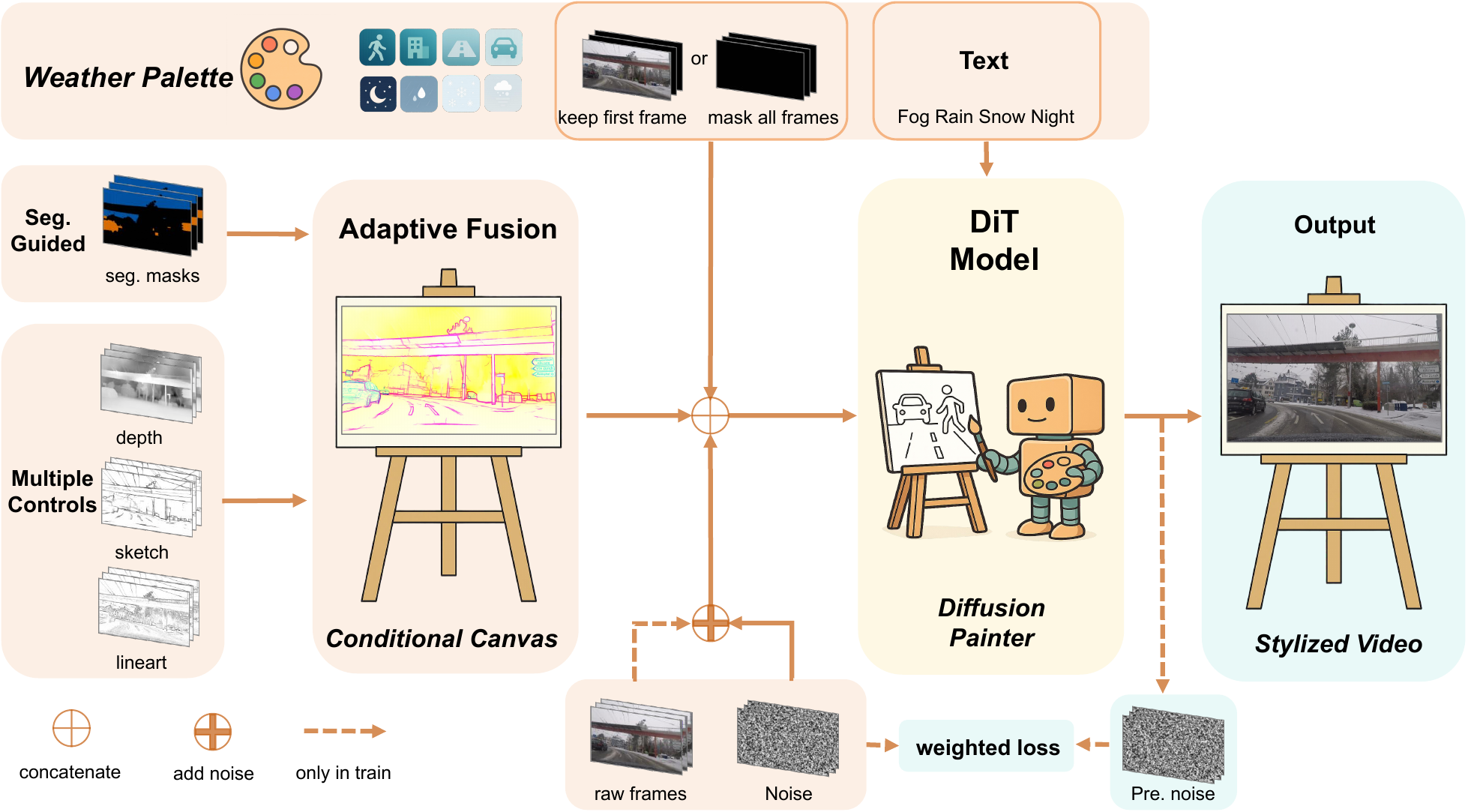}
  \caption{Overview of the proposed \emph{AutoAWG} for adverse weather generation.}
  \label{fig:framework}
\end{figure*}

\section{Related Work}
\label{sec:related}





\subsection{Adverse Weather in Autonomous Driving}

\noindent \textbf{Adverse Weather Removal.}
Many methods aim to enhance perception in adverse weather by restoring clear visual cues. Initial studies focused on single-condition image restoration (e.g., de-raining, de-snowing, de-hazing)~\citep{ni2021controlling, valanarasu2022transweather, ozdenizci2023restoring}, while more recent works extend to multi-condition video restoration~\citep{yang2023video, yang2024genuine}. These models are often trained on synthetic datasets such as Outdoor-Rain~\citep{li2019heavy}, RainDrop~\citep{qian2018attentive}, and Snow100K~\citep{liu2018desnownet}, though recent efforts are shifting toward real-world data~\citep{zhu2023learning}. A comprehensive review is provided in~\citep{xiao2024multiple}.

\noindent \textbf{Adverse Weather Generation.}
Another direction involves generating adverse weather data to augment training datasets and improve robustness under challenging conditions. Early methods used GANs to synthesize weather effects~\citep{kwak2021adverse, lan2024sustechgan, li2022weather}, but their instability has led to interest in rendering-based approaches~\citep{wang2022r, zhao2024revisiting}. 

{While recent methods~\citep{wen2024panacea,chen2024unimlvg,qian2025wedit} can generate driving videos under various weather conditions, they have notable limitations. Panacea~\citep{wen2024panacea} relies on an image-to-image translation model to convert the initial frame.
Whereas UniMLVG~\citep{chen2024unimlvg} requires pre-training on large-scale, web-crawled data with complex labeling. 
WeatherEdit~\citep{qian2025wedit} adopts a multi-stage pipeline with explicit 3D reconstruction, which increases complexity and makes scaling to long videos challenging. 
Moreover, none of these methods explicitly considers the preservation of scene details, making the generated videos unsuitable for direct reuse of existing annotations.}

\subsection{Video Generation in Autonomous Driving}

Recent studies on generation for autonomous driving can be broadly categorized into two paradigms: 

\textbf{Reconstruction-based methods} aim to regenerate the 3D driving scene using multi-view images, often aided by LiDAR data. These include techniques based on NeRF~\citep{mildenhall2021nerf} and 3D Gaussian Splatting~\citep{kerbl20233d}, such as~\citep{yan2024street, zhou2024drivinggaussian}, which reconstruct detailed dynamic environments from onboard sensors.
\textbf{Controllable generation methods}, on the other hand, utilize diffusion models guided by structured conditions such as camera trajectories, BEV maps, or textual prompts~\citep{gao2023magicdrive, wang2024drivedreamer, wang2024driving, wen2024panacea, zhou2024simgen, wu2024drivescape, zhao2024drivedreamer2, ni2025maskgwm}. Some methods integrate reconstruction and generation~\citep{ni2024recondreamer, zhao2024drivedreamer4d}.

Our work falls into the controllable generation category, focusing on transforming existing autonomous driving videos into adverse weather conditions. Unlike reconstruction-heavy editing pipelines, our approach directly leverages pixel-level controls extracted from the input videos and preserves semantic–structural integrity, making existing annotations reusable without re-labeling and enabling plug-and-play augmentation for downstream perception tasks.

\section{Method}
\label{sec:method}

\subsection{Architectural Overview}
\label{sec:method_overview}

We formulate adverse weather generation as a video style transfer problem within a controllable diffusion framework. As illustrated in Figure~\ref{fig:framework}, multiple complementary control conditions are adaptively fused according to semantic masks to construct a conditional canvas. The DiT model then acts as a “Painter,” filling this canvas with realistic weather effects while preserving object fidelity.

Given an input video, multiple structural controls are first extracted and safety-critical objects are segmented. During training, frames are encoded into the latent space via a 3D VAE, perturbed with Gaussian noise, and progressively denoised through the DiT. At inference, generation starts from random noise, guided by a target weather embedding and the fused conditions, to produce a temporally consistent transformed sequence. Detailed descriptions of each component are provided in the following sections.

\begin{figure}
  \centering
  \includegraphics[width=\linewidth]{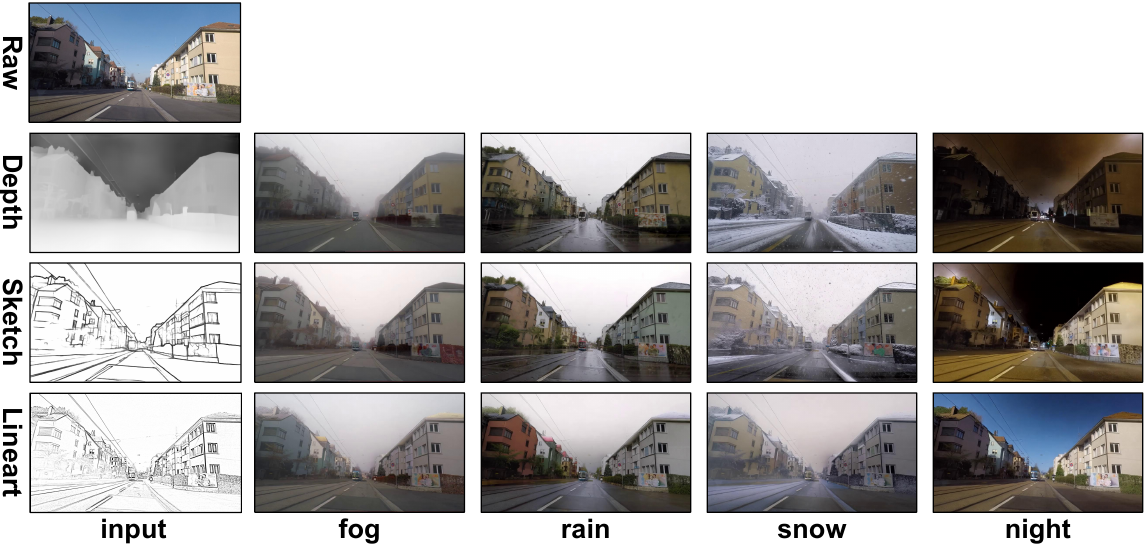}
  \caption{Comparison of different control maps and generated results.}
  \label{fig:control_strength}
\end{figure}

\begin{figure}
  \centering
  \includegraphics[width=\linewidth]{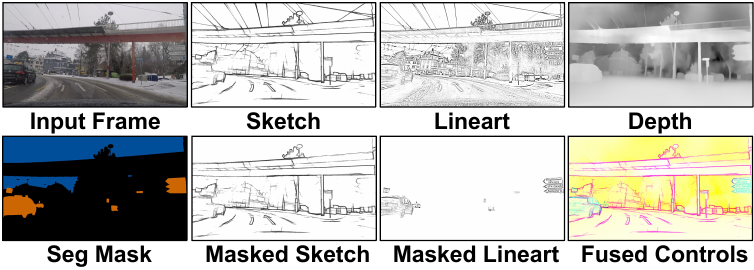}
  \caption{Adaptive fusion of controls guided by segmentation masks.}
  \label{fig:controls}
\end{figure}

\subsection{Adaptive Fusion of Multiple Controls}
\label{sec:method_mul_cons}

We propose a region-aware adaptive fusion strategy to integrate multiple control maps of varying strengths.

\paragraph{Control Strength Spectrum.}
We observe that different control conditions vary in how strongly they relate to the source content. Some control conditions capture a fine details, making it easy to generate content that closely resembles the original imagery but difficult to adapt to other styles. In contrast, others extract only coarse structural outlines, making it harder to reconstruct the original content details but facilitating strong style transformation. 

As shown in Figure~\ref{fig:control_strength}, the depth map~\citep{yin2021learning} captures only coarse structural layouts, which facilitates strong transformations into adverse weather conditions. In contrast, the lineart map retains fine-grained details, even down to textures such as tree bark, thereby constraining large stylistic changes. The sketch map~\citep{su2023lightweight} lies between these two extremes, achieving a balance between content preservation and style flexibility. This trend is further corroborated by the results in Table~\ref{tab:ablation}.

\paragraph{Region-Aware Control Fusion.}
Building on these findings, we selectively combine control maps using semantic masks. Critical objects (e.g., vehicles, pedestrians, traffic signs) must remain consistent, so only their regions are retained in the lineart map. Sky regions are removed from sketch maps to allow flexible weather variations, while depth maps are preserved entirely. The fused control is formulated as:
\begin{equation}
    \mathbf{C}_{a}=\text{concat}\{\mathbf{C}_{d}, \mathbf{C}_{l} \odot \mathbf{M}_{obj}, \mathbf{C}_{s} \odot (\mathbf{1} - \mathbf{M}_{sky}) \},
    \label{eq:control_combine}
\end{equation}
where $\mathbf{C}_{d}, \mathbf{C}_{l}, \mathbf{C}_{s}$ denote depth, lineart, and sketch maps, and $\mathbf{M}_{obj}, \mathbf{M}_{sky}$ are object and sky masks. Figure~\ref{fig:controls} illustrates this segmentation-aware multi-control fusion. Fused controls are then combined into a video and encoded via the 3D VAE.

\begin{figure}
  \centering
  \includegraphics[width=\linewidth]{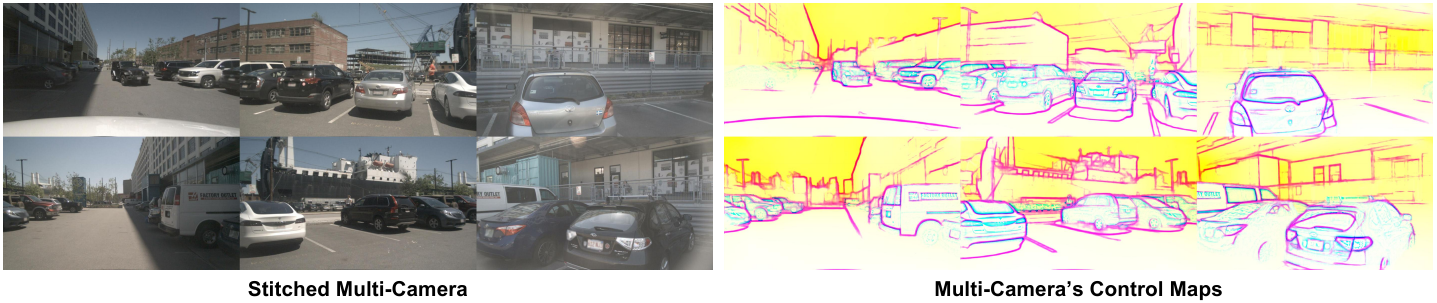}
  \caption{Multi-camera scenario: stitched frames and corresponding controls.}
  \label{fig:multi_view}
\end{figure}

\subsection{Importance-Weighted Loss}
\label{sec:method_lossfunc}

We adopt the Flow Matching framework~\citep{lipmanflow}. For a latent video sample $\mathbf{X}_1$ and Gaussian noise $\mathbf{X}_0 \sim \mathcal{N} (\mathbf{0}, \mathbf{1})$, the interpolated state is $\mathbf{X}_t = t \mathbf{X}_1 + (1-t) \mathbf{X}_0$. The model predicts the velocity $\mathbf{V}_t$ to approximate the ground truth $\mathbf{U}_t = d\mathbf{X}_t/dt$. Standard training minimizes:
\begin{equation}
    \mathcal{L} = \mathbb{E} {\left \| \mathbf{U}_t - \mathbf{V}_t \right \|}^2.
    \label{eq:mse_loss}
\end{equation}

However, uniform weighting overlooks varying regional importance. To emphasize critical regions, we introduce an importance-weighted loss:
\begin{equation}
    \mathcal{L} = \mathbb{E} \left [ {\left \| (\mathbf{U}_t - \mathbf{V}_t) \right \|}^2 + \alpha \cdot {\left \| \mathbf{M}_{obj} \odot (\mathbf{U}_t - \mathbf{V}_t) \right \|}^2 \right ],
    \label{eq:weighted_loss}
\end{equation}
where $\mathbf{M}_{obj}$ denotes object masks and $\alpha$ controls relative importance. This prioritizes accurate reconstruction of critical objects such as vehicles and pedestrians.

\subsection{Multi-View and Unlimited-Length Generation}
\label{sec:method_multiview}

Our framework naturally extends to multi-camera setups in autonomous driving. Since control maps operate at the pixel level, spatially aligned and temporally synchronized inputs ensure consistent generation across views. 
We stitch frames from all cameras into a single composite grid and apply the same operation to their corresponding control maps (Figure~\ref{fig:multi_view}), ensuring uniform appearance of shared objects across cameras. Compared to prior methods that rely on explicit cross-view attention~\citep{gao2023magicdrive, wen2024panacea}, our approach is simpler yet effective. 
The spatial concatenated multi-view frames are treated as a joint input, which is then partitioned into tokens within the DiT architecture. Through self-attention, DiT models the relationships among these tokens, enabling our approach to implicitly capture the correspondences and consistency across different camera views.

For unlimited-length sequences, we adopt a segment-wise strategy using specific inpainting masks. During training, we randomly mask either all frames (for generating the first segment) or all but the first frame (for continuation generation). At inference, each new segment is conditioned on its control maps and the last frame of the previous segment (none for the first segment), ensuring temporal continuity over arbitrarily long sequences.

\begin{figure}
  \centering
  \includegraphics[width=\linewidth]{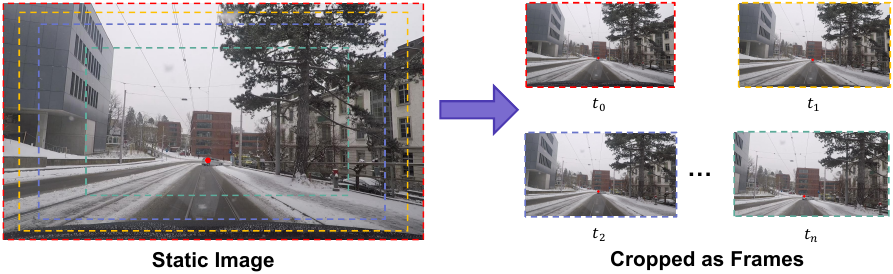}
  \caption{VP-Anchored Temporal Synthesis: synthesize video from a single image.}
  \label{fig:img2vid}
\end{figure}

\subsection{Adverse Weather Training Data}
\label{sec:method_training_data}

To compensate for the lack of open-source diverse adverse-weather driving videos, we construct a mixed training set from image and video datasets. Specifically, we propose a crop-to-video strategy to convert ACDC images to videos, and use them together with nuScenes samples to train our model.

\paragraph{ACDC Crop-to-video Dataset.} 
ACDC~\citep{sakaridis2021acdc} contains 4,006 images across fog, rain, snow, and nighttime. As it lacks temporal sequences, we propose the \textbf{VP(Vanishing Point)-Anchored Temporal Synthesis} strategy to synthesize videos from static images.
Specifically, we first estimate the vanishing point~\citep{pautrat2023vanishing} in the image and then continuously crop the image using a fixed aspect ratio, ensuring that the vanishing point's relative position remains unchanged in each cropped image. The original image resolution is 1920×1080, and the final cropped image is fixed at 960×544, with the resolutions of the intermediate images decreasing uniformly. Finally, all the cropped images are resized to the same resolution and concatenated along the temporal dimension to produce a video. After that, each image in ACDC is thus transformed into a 45-frame pseudo-video simulating driving motion. Figure~\ref{fig:img2vid} demonstrated this process.

\paragraph{nuScenes Dataset.} 
nuScenes~\citep{caesar2020nuscenes} is a leading benchmark for autonomous driving research which provides 1,000 scenes of 20-second multi-camera videos at 12 FPS, annotated with 3D bounding boxes. Although only nighttime and rainy conditions are included, its large-scale, multi-view setting makes it an ideal complement to ACDC-derived videos. Together, they form a comprehensive dataset for training adverse weather generation.

\section{Experiments}
\label{sec:exp}

\subsection{Evaluation Metrics}
\label{sec:exp_metric}

\begin{table}
\centering
    \caption{Quantitative comparison with automotive video generation methods on nuScenes dataset.}
    \label{tab:compare}
    \centering
    \begin{tabular}{c|c|cc}
    \toprule
    Methods & w/ 1st frame & FID$\downarrow$ & FVD$\downarrow$ \\\midrule
    MagicDrive~\citep{gao2023magicdrive} & $\times$ & - & 217.9 \\
    MagicDrive-V2~\citep{gao2024magicdrivedit} & $\times$ & - & 94.8 \\
    DriveDreamer~\citep{wang2024drivedreamer} & $\times$ & 26.8 & 353.2 \\
    DriveDreamer-2~\citep{zhao2024drivedreamer2} & $\times$ & 25.0 & 105.1 \\
    DiVE~\citep{jiang2024dive} & $\times$ & - & {94.6} \\
    Ours & $\times$ & \textbf{12.5} &  \textbf{79.4}  \\
    \midrule
    GenAD~\citep{yang2024generalized} & $\checkmark$ & 15.4 & 244  \\
    Drive-WM~\citep{wang2024driving} & $\checkmark$ & 15.8 & 122.7  \\
    Panacea~\citep{wen2024panacea} & $\checkmark$ & 16.9 & 139.0 \\
    Vista~\citep{gao2024vista} & $\checkmark$ & {6.9} & 89.4 \\
    DriveDreamer-2~\citep{zhao2024drivedreamer2} & $\checkmark$ & 11.2 & {55.7} \\
    GEM~\citep{hassan2025gem} & $\checkmark$ & 10.5 & 158.5 \\
    Glad~\citep{XieLWC025} & $\checkmark$ & 11.2 & 188.0 \\
    {DriveScape~\citep{wu2024drivescape}} & {$\checkmark$} & {8.3} & {76.4} \\
    {MaskGWM~\citep{ni2025maskgwm}} & {$\checkmark$} & {8.9} & {65.4} \\
    Ours & $\checkmark$ & \textbf{6.3}  & \textbf{51.7}  \\    \bottomrule
    \end{tabular}
\end{table}

We evaluate our method from two aspects: the success of weather generation and the preservation of original scene content. For the weather alignment, we use CLIP~\citep{radford2021learning} to classify each generated frame according to weather (sunny, rainy, foggy, snowy) or time (daytime or nighttime). The average classification accuracy is reported as \emph{Weather} Score.
To asses the content preservation, we employ the widely used FID~\citep{heusel2017gans} and FVD~\citep{unterthiner2018towards}.
Additionally, we use mAP from object detection like several prior works~\citep{gao2023magicdrive}. 
Specifically, we apply the YOLO11X~\footnote{\url{https://docs.ultralytics.com/models/yolo11/}} to detect traffic elements in the transformed frames, and compute mAP using the COCO evaluation protocol~\citep{lin2014microsoft}.

\subsection{Implementation Details}
\label{sec:exp_impl}

We train our model using 8 NVIDIA H20 GPUs. All experiments were conducted in a 45-frame configuration, with the single-camera resolution fixed at 960×544. Firstly, we use the nuScenes dataset for initial training to capture the patterns of autonomous driving scenes. Subsequently, we finetune the model on a combined dataset consisting of nuScenes and the ACDC synthesized video, so that it can learn adverse weather effects. For multi-camera scenarios, we further finetune the model on nuScenes 6-camera videos.

For the ACDC dataset, we use the annotations for ``human", ``vehicle", ``traffic light", ``traffic sign" to construct the critical object mask $\mathbf{M}_{obj}$, and the ``sky" to build the sky mask $\mathbf{M}_{sky}$. For frames in the nuScenes dataset without segmentation masks, we use DeepLabv3~\citep{chen2017rethinking} pretrained on the Cityscapes dataset~\citep{cordts2016cityscapes} to obtain these masks. We use the controlnet-aux toolbox to extract all the control conditions.

We adopt CogVideoX1.5-5B~\citep{yang2024cogvideox} as our backbone, and use its 3D VAE to encode the fused control maps, rather than training an encoder from scratch.

\subsection{Quantitative Evaluation}
\label{sec:exp_compare}

\paragraph{Generation Quality.}
To assess the visual quality of generated automotive videos, we compare our method against several state-of-the-art approaches in the field of autonomous driving video generation. Following the evaluation protocols in prior works~\citep{wang2024drivedreamer, wang2024driving, zhao2024drivedreamer2}, we conduct quantitative analysis on the nuScenes validation set.
To ensure a fair comparison, we configure our model to transform each input video into the same weather condition as its original, rather than into an adverse condition. This setup allows us to focus purely on evaluating the generative quality of the scene. 
As shown in Table~\ref{tab:compare}, our method achieves an FID of 12.5 and FVD of 79.4 without the first-frame input, and further improves to an FID of 6.3 and FVD of 51.7 when conditioned on the first frame, both substantially surpassing previous state-of-the-art methods. These results highlight the strong ability of our framework to generate high-quality automotive videos.

\paragraph{Downstream Utility Evaluation.}  
To evaluate the practical value of our generated videos, we examine their effectiveness in enhancing downstream perception tasks. Specifically, we augment the nuScenes training set with our generated frames (limited to sunny, rainy, and nighttime scenes to match the validation distribution), and use this combined dataset to train a camera-only BEVFusion model~\citep{liu2023bevfusion} for 3D object detection.
As shown in Table~\ref{tab:effect_gen}, incorporating our synthetic data leads to noticeable improvements: the model's mAP increases by 1.99 points, and the NDS improves by 1.36 points. These results indicate that our generated videos not only exhibit high visual quality but also offer tangible benefits for real-world perception algorithms in autonomous driving.

\paragraph{Edit Fidelity.}  
We evaluate edit fidelity following \citet{gao2024magicdrivedit} by transferring validation cases to normal, rainy, and nighttime conditions while retaining the same ground-truth annotations as raw frames. The pretrained BEVFusion~\citep{liu2023bevfusion} is then applied for object detection. As shown in Table~\ref{tab:controllability}, the detection mAP of our generated frames remains much closer to that of raw frames, whereas MagicDrive-V2~\citep{gao2024magicdrivedit} exhibits larger drops. These results demonstrate that our method more effectively preserves key scene elements across weather transformations.

\begin{table}
  \centering
    \caption{Impact of generated data on BEVFusion 3D object detection on nuScenes dataset.}
    \label{tab:effect_gen}
    \centering
    \begin{tabular}{c|cc}
    \toprule
    Methods & mAP$\uparrow$ & NDS$\uparrow$ \\
    \midrule
    w/o gen. data & 35.53 & 41.20 \\
    w/ gen. data & \textbf{37.52}\textcolor{red}{\scriptsize{+1.99}} & \textbf{42.56}\textcolor{red}{\scriptsize{+1.36}} \\
    \bottomrule
    \end{tabular}
\end{table}

\begin{table}
\centering
    \caption{Comparison with MagicDrive-V2 for controllable generation. mAP on origin and generated frames are shown.}
    \label{tab:controllability}
    \centering
    \begin{tabular}{c|c|c|c}
    \toprule
    {data split} & {method} & {mAP}$\uparrow$ & {mAP drop}$\downarrow$ \\
    \midrule
    \multirow{3}{*}{all} & Raw & 0.3553 & - \\ 
    & MagicDrive-V2 & 0.1817 & 0.1736 \\ 
    & Ours & 0.3376 & 0.0177 \\ 
    \midrule
    \multirow{2}{*}{rainy} & Raw & 0.3435 & - \\ 
    & Ours & 0.2931 & 0.0504 \\ 
    \midrule
    \multirow{2}{*}{night} & Raw & 0.1801 & - \\ 
    & Ours & 0.1645 & 0.0156 \\ 
    \bottomrule
    \end{tabular}
\end{table}

\subsection{Qualitative Evaluation}
\label{sec:exp_qualitative}

\begin{figure}
\centering
    \begin{subfigure}{\linewidth}
    \centering
    \includegraphics[width=\linewidth]{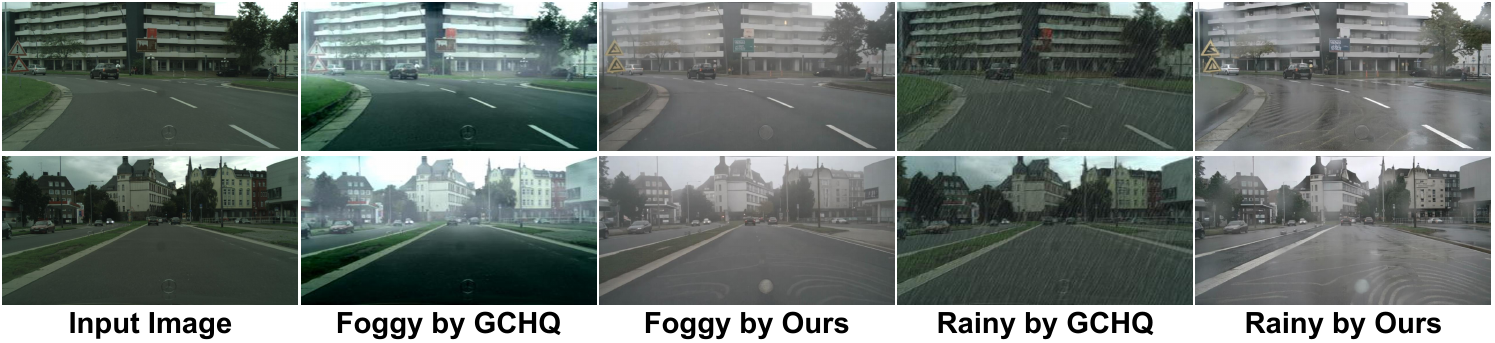}
    \caption{
        Visual comparison with GCHQ~\citep{zhao2024revisiting}.
    }
  \label{fig:vs_weather:gchq}
  \end{subfigure}
  \begin{subfigure}{\linewidth}
      \centering
        \includegraphics[width=\linewidth]{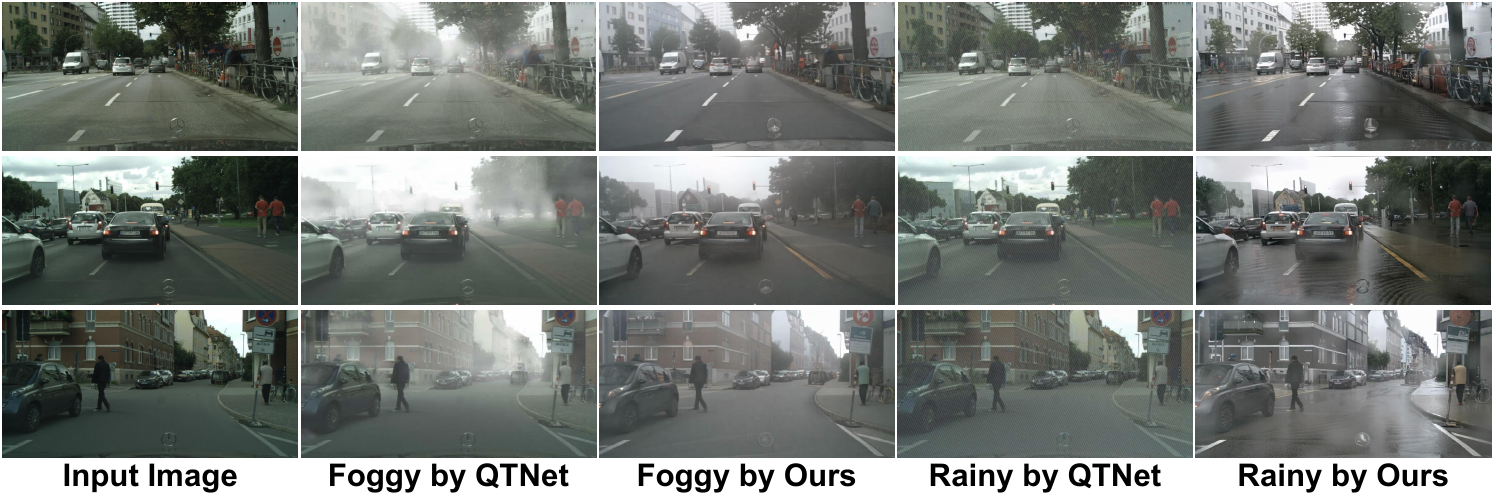}
        \caption{
          Visual comparison with QTNet~\citep{wang2022r}.
          }
        \label{fig:vs_weather:qtnet}
  \end{subfigure}
\caption{Visual comparison with image-based weather translation for foggy and rainy weather translation. Our method achieves comparable or better weather effects without dataset-specific training.}
\label{fig:vs_weather}
\end{figure}

\paragraph{Comparison with Image-based Weather Translation.}  
We first compare our method with existing image-based weather transformation models.
Figure~\ref{fig:vs_weather} shows the foggy and rainy results generated by GCHQ~\citep{zhao2024revisiting}, QTNet~\citep{wang2022r} and our approach using the same input images, which are chosen from Cityscapes~\citep{cordts2016cityscapes}. 
{Despite not being trained on this dataset, our model still produces realistic and high-quality adverse weather effects}, demonstrating comparable or even superior visual fidelity to GCHQ and QTNet. {This highlights the strong generalization ability of our model.}

\begin{figure}
  \centering
  \includegraphics[width=\linewidth]{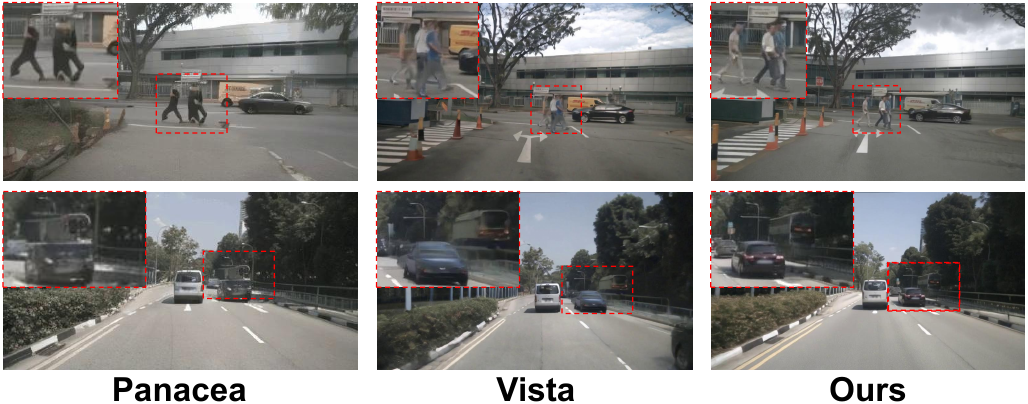}
  \caption{Visual comparison with Panacea~\citep{wen2024panacea} and Vista~\citep{gao2024vista}. Our approach generates more realistic and detailed vehicles, pedestrians, and obstacles compared to Panacea and Vista.}
  \label{fig:vs_sota}
\end{figure}

\paragraph{Comparison with Video-based Generation.} 
To visually compare our method with state-of-the-art approaches, we generate the same nuScenes cases using our model and two leading methods: Vista~\citep{gao2024vista} and Panacea~\citep{wen2024panacea}. Results are shown in Figure~\ref{fig:vs_sota}.
Our method produces more realistic traffic elements, including vehicles, pedestrians, traffic cones, and lane markings. In comparison, the outputs from Vista and Panacea exhibit noticeable artifacts or less accurate structural details.

\begin{figure}
  \centering
  \includegraphics[width=\linewidth]{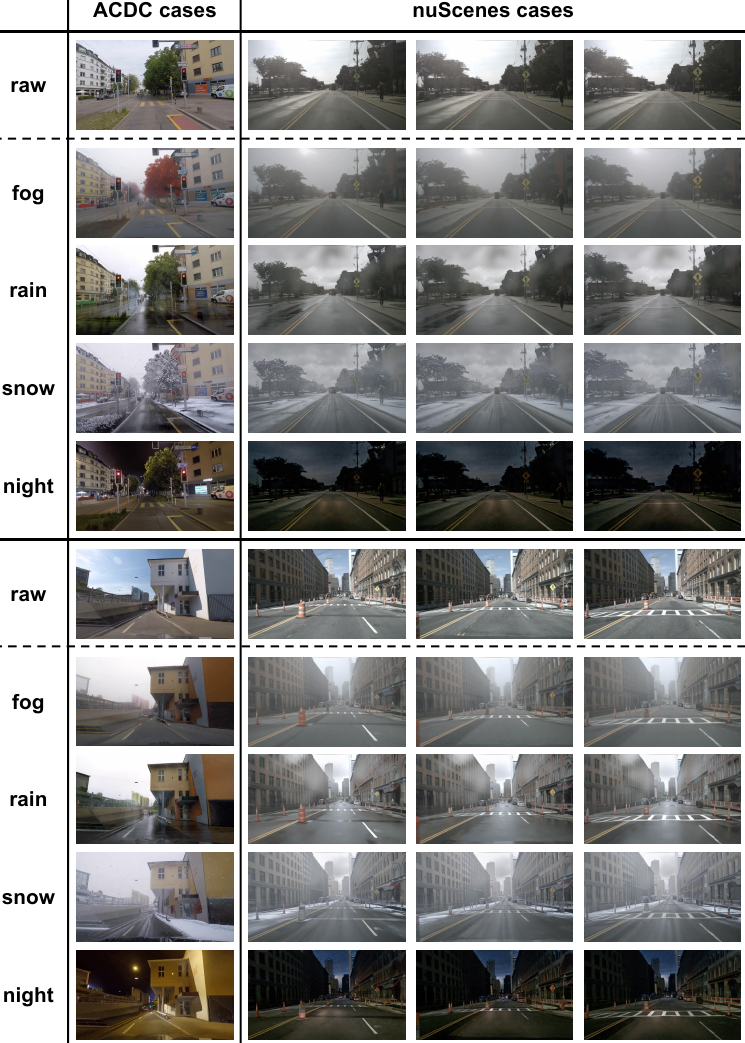}
  \caption{
  Adverse weather transformation results for ACDC images and nuScenes key frames.
  }
  \label{fig:transfer_results}
\end{figure}

\paragraph{Controllability of Weather.} 
To show the controllability of our method in generating adverse weathers. Our model is used to transform normal videos to four adverse conditions: foggy, rainy, snowy, and nighttime.
Results are shown in Figure~\ref{fig:transfer_results}. For ACDC cases, we show the first generated frame. For nuScenes videos, we show three representative frames to highlight temporal consistency. From it we can see, our method is capable of producing visually realistic and temporally coherent weather effects while preserving the scene structure—such as vehicles, pedestrians, traffic signs and lights.

Notably, even though the nuScenes dataset does not contain any foggy or snowy scenes, our model successfully synthesizes realistic foggy and snowy effects for nuScenes videos, thanks to its generalization ability learned from the ACDC synthesized videos.

\begin{figure}
  \centering
  \includegraphics[width=\linewidth]{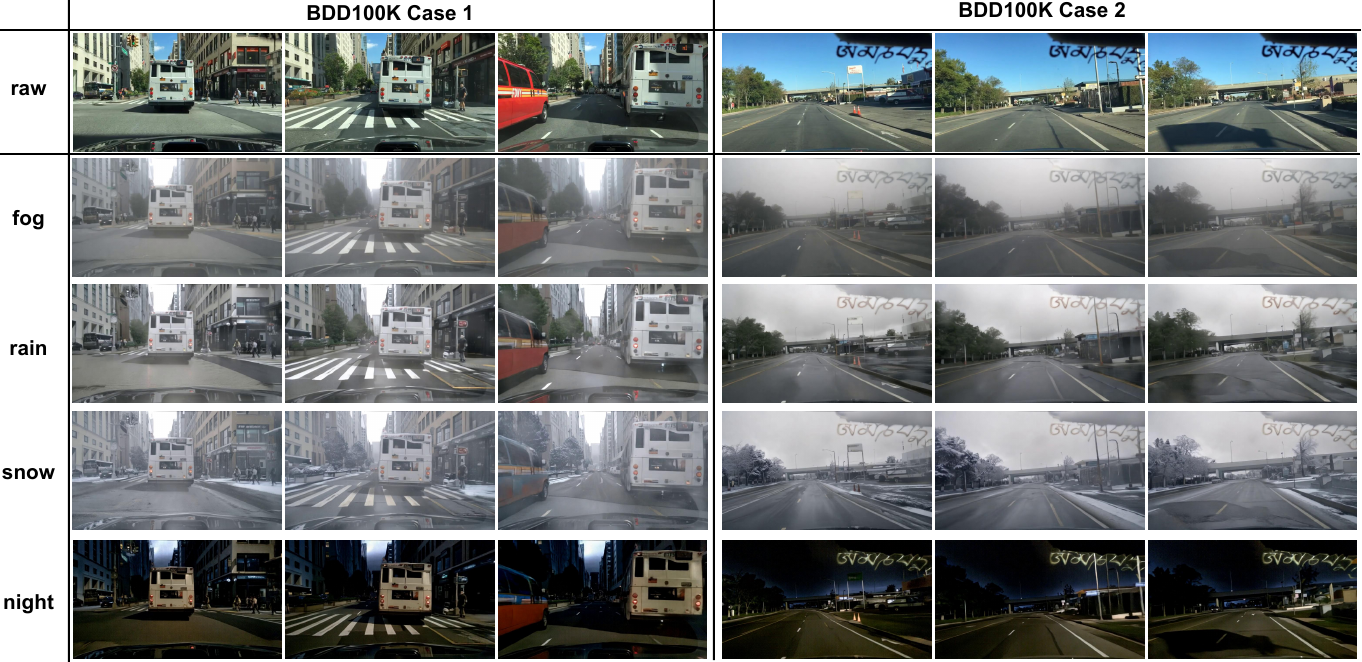}
  \caption{{Transferring BDD100K samples to adverse weather using AutoAWG.}}
  \label{fig_app:bdd100k}
\end{figure}

\begin{figure*}
  \centering
  \includegraphics[width=\linewidth]{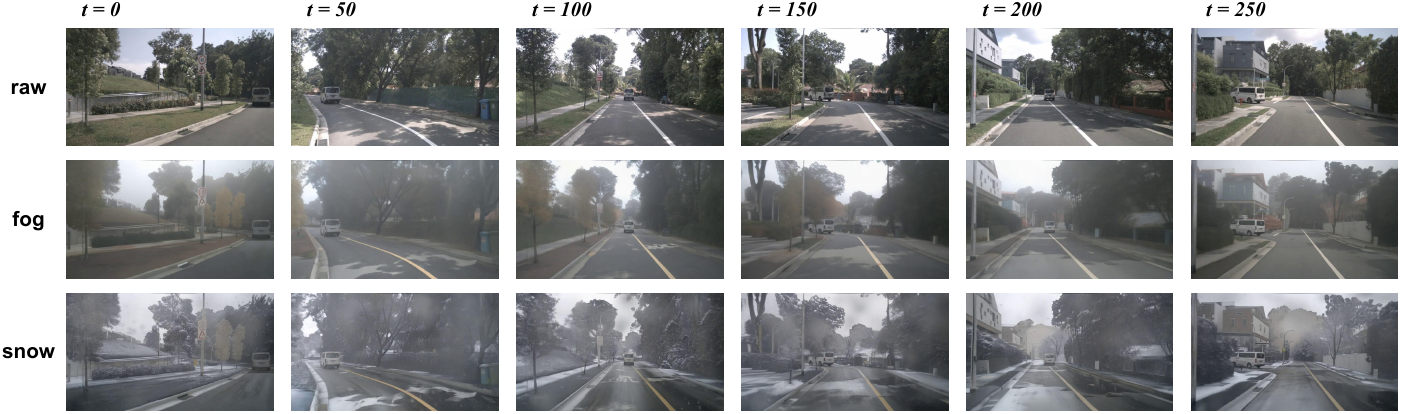}
  \caption{Long video generation results. The results show limited degradation over time, and the white van remains consistent.}
  \label{fig:long_video}
\end{figure*}

\paragraph{Generalization to Other Datasets.} 
Despite being trained on only a few thousand ACDC images and several hundred nuScenes video clips, our AutoAWG demonstrates strong generalization to unseen datasets. As shown in Figures~\ref{fig:vs_weather}, the model effectively transfers Cityscapes images to multiple adverse weather conditions. We further validate AutoAWG on the BDD100K dataset~\citep{bdd100k}, and Figure~\ref{fig_app:bdd100k} illustrates its ability to transform BDD100K video frames as well. These results collectively indicate that AutoAWG achieves robust and reliable cross-dataset generalization, even under limited training data.

\begin{figure}
  \centering
  \includegraphics[width=\linewidth]{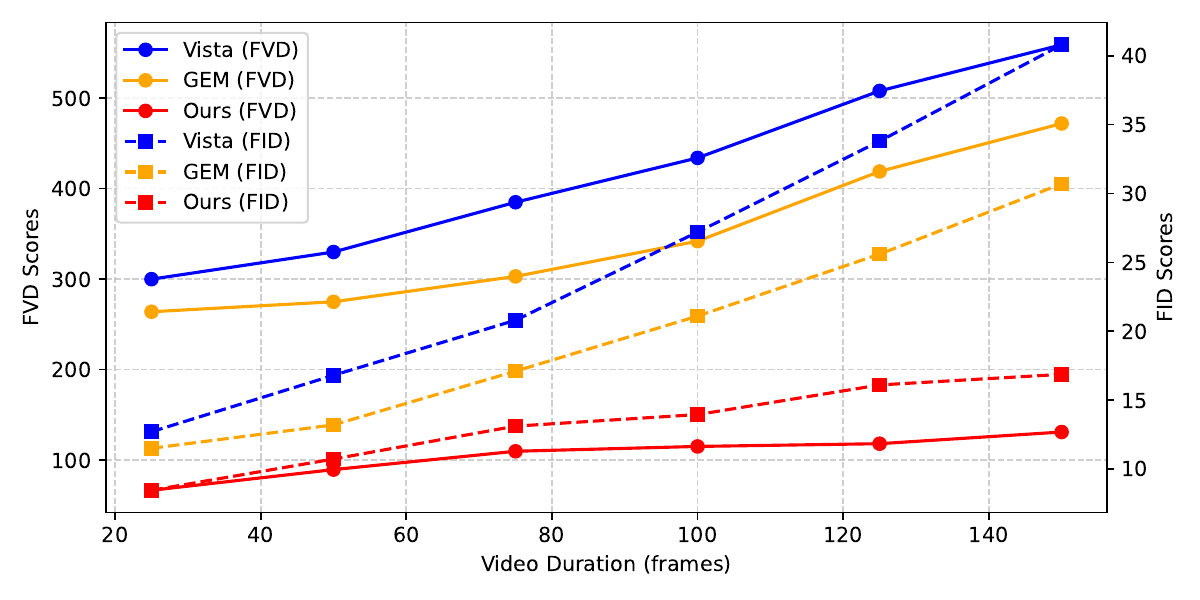}
  \caption{FID and FVD for long video generations. Our method achieves lower and more stable scores than GEM~\citep{hassan2025gem} and Vista~\citep{gao2024vista}.}
  \label{fig:long_stability}
\end{figure}

\subsection{Multi-Camera and Long Video Generation}
\label{sec:exp_multiview}

To assess both cross-view consistency and long-horizon stability, we train a 6-camera model on the nuScenes dataset and apply it to sequences exceeding 250 frames. 

Figure~\ref{fig:long_video} presents long-sequence generation, with raw and translated frames at six sampled timestamps. The results reveal no perceptible degradation in visual quality over extended durations. Notably, the white van remains coherent and well-preserved across all 250 frames, demonstrating robust temporal consistency.

We further quantify long-horizon stability following~\citet{hassan2025gem}, computing FID and FVD on subsequences of 25, 50, 75, 100, 125, and 150 frames. As shown in Figure~\ref{fig:long_stability}, our method significantly outperforms Vista~\citep{gao2024vista} and GEM~\citep{hassan2025gem}. Specifically, the score drop between the first and last segments is only 8.4 (FID) and 64.4 (FVD), compared to 19.2 \& 208 for GEM and 28.1 \& 259 for Vista. These results highlight the superior temporal consistency and overall quality of our approach, even for long video sequences.

\begin{figure}
  \centering
  \includegraphics[width=\linewidth]{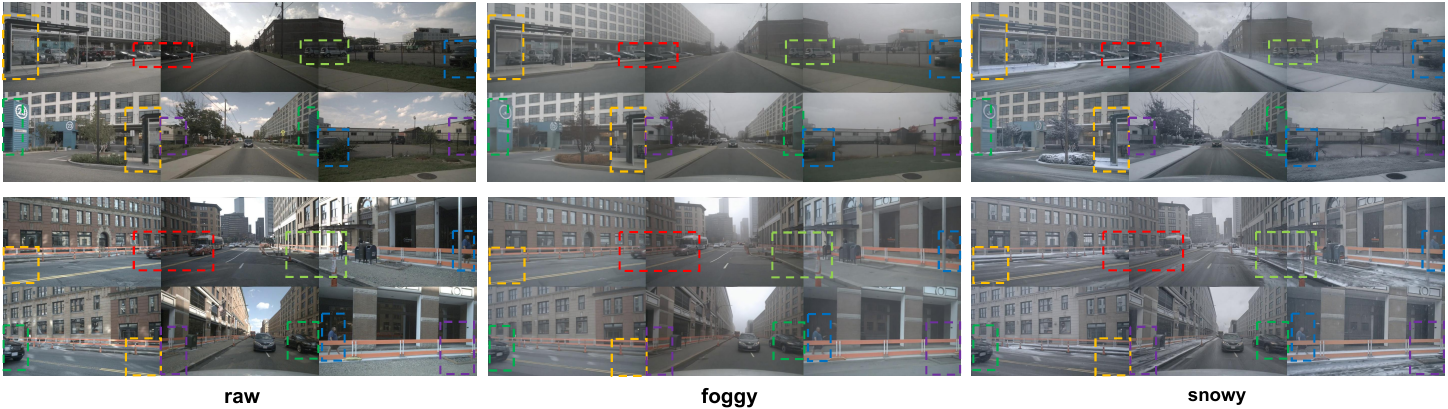}
  \caption{Multi-camera results on nuScenes. The dashed boxes in same color indicate same objects in each cameras.}
  \label{fig:nuscene_6v}
\end{figure}

Representative results within multi-camera are shown in Figure~\ref{fig:nuscene_6v}, where six synchronized raw camera views and their foggy and snowy translations are displayed. Objects appearing across different cameras remain visually consistent, underscoring the model’s ability to maintain strong cross-camera coherence.

\begin{figure*}
  \centering
  \includegraphics[width=\linewidth]{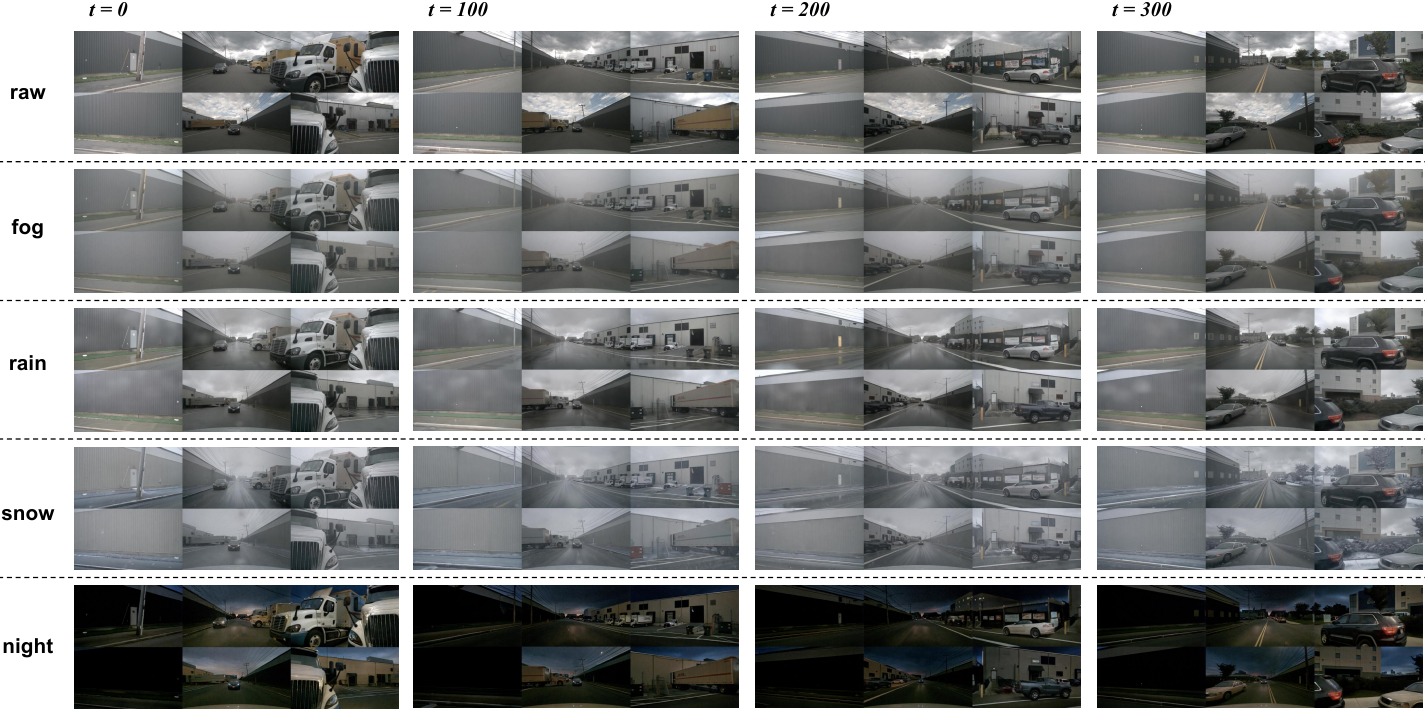}
  \caption{6-view long-duration weather transformation results on nuScenes.}
  \label{fig:nuscene_6v_1}
\end{figure*}

Figures~\ref{fig:nuscene_6v_1} present additional results of our method on multi-camera, long-duration weather transformation. Each input sequence is converted into four adverse conditions: fog, rain, snow, and nighttime. For each weather type, we display representative frames sampled at the 0th, 100th, 200th, and 300th frames to illustrate the temporal progression.

\subsection{Ablation Study}
\label{sec:exp_ablation}

We conduct ablation studies on ACDC samples, each translated into four adverse conditions: fog, rain, snow, and nighttime. The results, summarized in Table~\ref{tab:ablation}, highlight the distinct control strengths of different inputs: depth and sketch enhance weather realism but reduce content fidelity, while lineart shows the opposite trend, consistent with the visual analysis in Figure~\ref{fig:control_strength}.

For multi-control fusion, we compare four strategies: (1) separate encoding, (2) concatenated encoding, (3) mask-based fusion with separate encoding, and (4) mask-based fusion with concatenated encoding. Simple separate or concatenated fusion preserves objects well but yields limited weather realism, resembling lineart control. By contrast, our mask-based fusion strikes a better balance between structure and realism. In particular, the ``masked-concat'' strategy matches the performance of ``masked-separate'' while being more efficient, making it the preferred choice. Finally, adding the importance-weighted loss {(the \emph{AutoAWG} row)} further improves detection accuracy while maintaining visual realism, confirming its effectiveness in enhancing consistency of key objects.

\begin{table}
\centering
    \caption{Ablation study on our method.}
    \label{tab:ablation}
    \begin{tabular}{c|c|c}
    \toprule
    Methods & Weather$\uparrow$ & mAP$\uparrow$    \\ \midrule
    depth control & 1  & 0.5248 \\
    sketch control  & 1  & 0.5326 \\
    lineart control & 0.5611  & 0.6516 \\ 
    \midrule
    separate 3 controls  & 0.5328  & 0.6784 \\
    concat 3 controls  & 0.5444  & 0.6672 \\
    masked separate & 0.9586 & 0.6325 \\
    masked concat  & 0.9567  & 0.6394 \\ 
    \midrule
    \emph{AutoAWG}  & 0.9506  & 0.6594 \\
    \bottomrule
    \end{tabular}
\end{table}

Also, we perform an ablation study to determine the optimal value of the additional loss weight $\alpha$ applied to critical regions. Following the same setup as in the main ablation study, we transform 20 ACDC crop-to-video samples into four adverse conditions, and vary $\alpha$ among \{0.5, 1.0, 1.5, 2.0\} during training. The results, summarized in Table~\ref{tab_app:ablation}, show that $\alpha = 1.0$ achieves the best balance between weather realism (Weather Score) and content fidelity (Detection mAP), and is therefore used as the default setting.

\begin{table}
\centering
  \caption{Effect of varying the loss weight $\alpha$ on transformation effects.}
  \label{tab_app:ablation}
    \begin{tabular}{c|c|c}
    \toprule
    $\alpha$ & Weather Score$\uparrow$ & Detection mAP$\uparrow$    \\ 
    \midrule
    0.5  & 0.9508  & 0.6551 \\
    1.0  & 0.9506  & 0.6594 \\
    1.5  & 0.9394  & 0.6628 \\
    2.0  & 0.9139  & 0.6575 \\
    \bottomrule
    \end{tabular}
\end{table}

\section{Conclusion}
\label{sec:conclusion}

This paper presents \emph{AutoAWG}, a novel framework for generating adverse weather effects in automotive videos, featuring an adaptive multi-control selection mechanism and an importance-weighted loss strategy. 
Our method effectively balances the realism of weather transformation with the preservation of critical scene elements. 
By extracting control maps directly from the input video, the framework naturally extends to multi-camera configurations—commonly used in autonomous driving—ensuring robust spatiotemporal consistency across synchronized views.
Furthermore, our segment-wise processing strategy supports open-loop generation of videos with arbitrary lengths, enabling consistent long-duration transformation through progressive conditioning on pre-generated segments.
Extensive experiments demonstrate the framework's ability to produce visually realistic, semantically consistent, and temporally coherent transformations under various adverse weather conditions, highlighting its scalability and practical potential for autonomous driving simulation and perception enhancement.


\bibliographystyle{ACM-Reference-Format}
\bibliography{sample-base}

\end{document}